\documentclass[lettersize,journal]{IEEEtran}
\usepackage{amsmath,amsfonts}
\usepackage{algorithmic}
\usepackage{array}
\usepackage[caption=false,font=normalsize,labelfont=sf,textfont=sf]{subfig}
\usepackage{textcomp}
\usepackage{stfloats}
\usepackage{verbatim}
\usepackage{graphicx}

\usepackage{times}  
\usepackage{helvet}  
\usepackage{courier}  
\usepackage[hyphens]{url}  
\usepackage{mathrsfs}
\usepackage{tcolorbox}
\usepackage{enumitem}    
\usepackage{pifont}
\usepackage{listings}
\usepackage{arydshln}
\usepackage[misc]{ifsym}
\usepackage{pifont}
\usepackage{booktabs}
\usepackage{cite}

\def\BibTeX{{\rm B\kern-.05em{\sc i\kern-.025em b}\kern-.08em
    T\kern-.1667em\lower.7ex\hbox{E}\kern-.125emX}}
\usepackage{balance}
\begin{document}
\title{Multi-View Large Reconstruction Model via Geometry-Aware Positional Encoding and Attention}
\author{Mengfei Li, Xiaoxiao Long, Yixun Liang, Weiyu Li, Yuan Liu, Peng Li, \\Wenhan Luo, Wenping Wang, \textit{Fellow, IEEE}, and Yike Guo, \textit{Fellow, IEEE}
\thanks{Manuscript created November, 2024. The research was supported by Theme-based Research Scheme (T45-205/21-N) from Hong Kong RGC, and Generative AI Research and Development Centre from InnoHK. (Corresponding authors: Yike Guo, Xiaoxiao Long.)

M. Li, X. Long, Y. Liang, W. Li, P. Li, W. Luo and Y. Guo are with the Hong Kong University of Science and Technology, Hong Kong, China (e-mail: mliek@connect.ust.hk; xxlong@connect.hku.hk; lyxun2000@gmail.com; weiyuli.cn@gmail.com; plibp@conncet.ust.hk; whluo@ust.hk; yikeguo@ust.hk).

Y. Liu and W. Wang are with the University of Hong Kong, Hong Kong, China (e-mail: yuanly@connect.hku.hk; wenping@cs.hku.hk)
}
}

\markboth{IEEE TRANSACTIONS ON VISUALIZATION AND COMPUTER GRAPHICS,~Vol.~XX, No.~XX, XX~2020}%
{How to Use the IEEEtran \LaTeX \ Templates}

\maketitle

\begin{figure*}[ht]
\includegraphics[width=\textwidth]{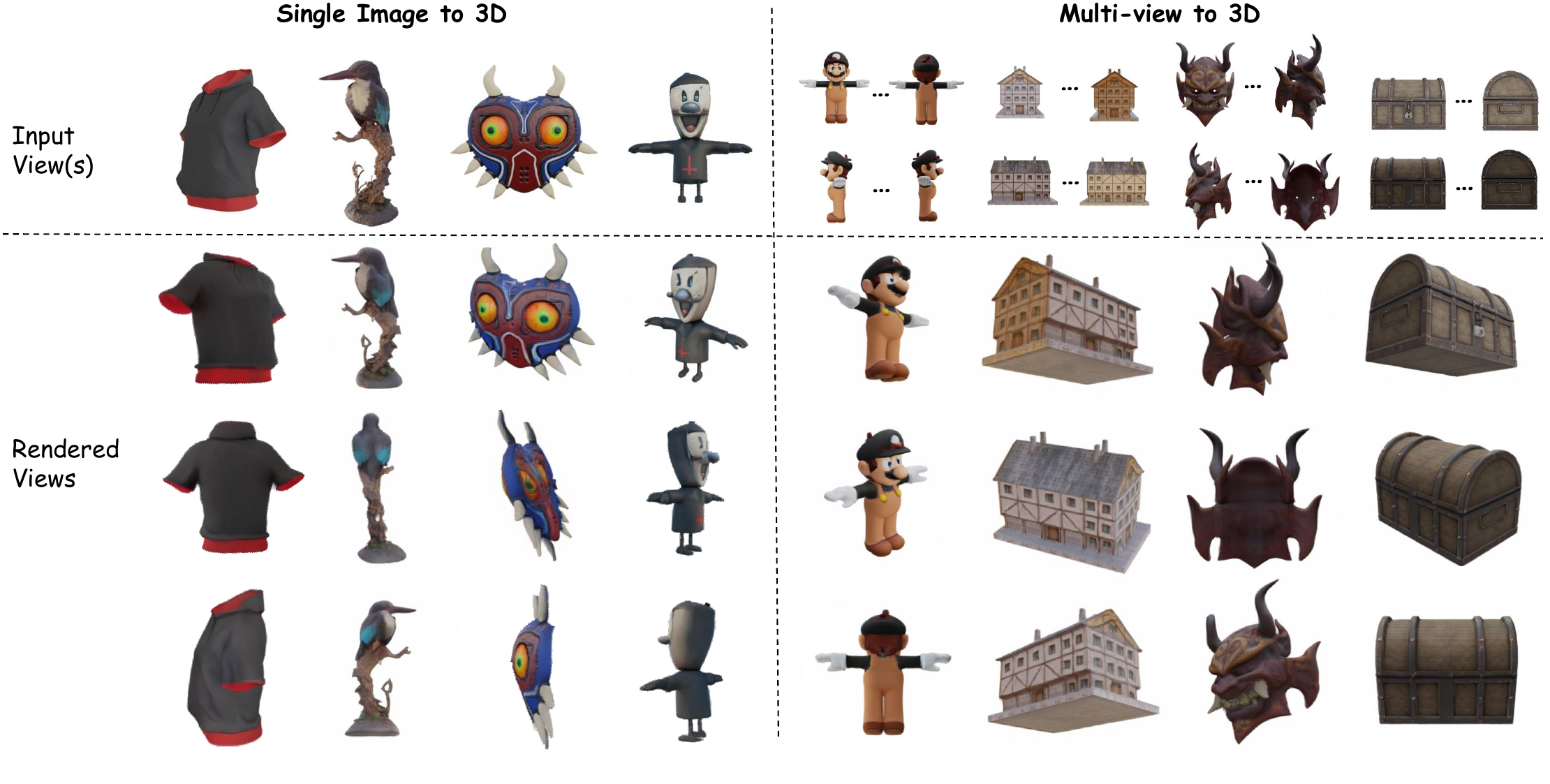}
\caption{Our M-LRM is a novel framework that can handle both single-view generation tasks and multi-view (e.g. 6-view) reconstruction tasks. M-LRM facilitates high-quality novel views generation given a single view or multiple views as input.}
\end{figure*}

\begin{abstract}
    Despite recent advancements in the Large Reconstruction Model (LRM) demonstrating impressive results, when extending its input from single image to multiple images, it exhibits inefficiencies, subpar geometric and texture quality, as well as slower convergence speed than expected.  
    It is attributed to that, LRM formulates 3D reconstruction as a naive images-to-3D translation problem, ignoring the strong 3D coherence among the input images.  In this paper, we propose a Multi-view Large Reconstruction Model (M-LRM) designed to reconstruct high-quality 3D shapes from multi-views in a 3D-aware manner. Specifically, we introduce a multi-view consistent cross-attention scheme to enable M-LRM to accurately query information from the input images. Moreover, we employ the 3D priors of the input multi-view images to initialize the triplane tokens. Compared to previous methods, the proposed M-LRM can generate 3D shapes of high fidelity. Experimental studies demonstrate that our model achieves a significant performance gain and faster training convergence. Project page: \url{https://murphylmf.github.io/M-LRM/}.
\end{abstract}

\begin{IEEEkeywords}
3D generation, large reconstruction model, sparse-view reconstruction, single image to 3D.
\end{IEEEkeywords}

\section{Introduction}
\IEEEPARstart{3D} RECONSTRUCTION from multi-view images stands as a fundamental and critical task that has garnered decades of study, playing a pivotal role in various downstream applications including virtual reality, 3D content generation, and robotic grasping. Achieving complete and high-quality reconstructions typically necessitates capturing images from dense viewpoints that offer comprehensive coverage of the target object.
However, traditional approaches relying on dense viewpoint images face significant degradation when confronted with limited inputs, such as few or even single images. This limitation results in distorted and incomplete reconstructions due to the inadequacy of information provided. Despite attempts to address 3D reconstruction from a few images or even a single image, existing methods often suffer from inefficiency or yield low-quality results.

Recently, the introduction of the Large Reconstruction Model (LRM) and its multi-view variant, Instant3D, has demonstrated remarkable results, indicating the feasibility of learning a generic 3D prior for reconstructing objects from single images or limited image sets. However, LRM approaches the task of 3D reconstruction from images by simplistically treating it as an image-to-3D regression problem. This involves directly training a highly scalable transformer on extensive data to learn the image-to-3D mapping.
Regardless of whether a single image or multiple images are provided as input, LRM first unpacks feature maps of the input images into tokens and then feeds the tokens into the transformer to inherently learn 3D priors. 
Consequently, when extending the input from a single image to multiple images, the LRM model fails to exhibit significant performance improvements since it disregards the 3D coherence among the input images.

In this paper, we introduce a novel architecture designed for reconstructing 3D shapes from multiple images in a 3D-aware manner, termed as the Multi-view Large Reconstruction Model (M-LRM). In contrast to the LRM, which primarily relies on regressing 3D shapes from images without considering their interrelations, our model draws inspiration from traditional Multi-view Stereo (MVS) techniques and explicitly learns correspondences among input images to infer the 3D geometry.
Our model builds upon the triplane-based transformer architecture introduced by LRM, leveraging its scalability on extensive datasets. Specifically, we propose a new multi-view consistent cross-attention scheme, which empowers M-LRM to infer 3D geometry by meticulously examining semantic consistency across input images. Furthermore, we introduce a geometry-aware positional embedding approach that directly incorporates the 3D information encoded in multiple images into the transformer architecture.
These two innovations enable M-LRM to learn 3D representations from images in a 3D-aware manner, thereby producing high-quality geometries.

Extensive experiments demonstrate that our M-LRM can generate high-quality 3D objects. It is capable of producing accurate views and geometry. Additionally, our experiments show that incorporating explicit 3D priors into the LRM model significantly speeds up convergence. In summary, our key contributions are as follows:
\begin{itemize}
    \item Our M-LRM addresses the shortcomings of instant3D, which lacks modeling of spatial correlation among multiple input views, leading to slow convergence and sub-optimal generation results. Instead, our M-LRM leverages the spatial coherence of input images to generate high-fidelity results.
    \item We propose a novel Geometry-aware Positional Embedding mechanism to fully utilize image prior. This helps the randomly initialized triplane capture spatial and image priors, enabling faster convergence of our M-LRM.
    \item We introduce a multi-view consistent cross-attention scheme to accurately query information from relevant image regions. This incorporates explicit spatial priors, making the 3D generative model geometry aware. As a result, our M-LRM can generate consistent novel views and high-quality geometry.
    \item Extensive experiments demonstrate that our M-LRM outperforms all baseline methods. It can also generate high-quality 3D objects with a single image input when equipped with a multi-view generation model like Zero123++~\cite{shi2023zero123++}.
    
\end{itemize}

\section{Related Work}


\subsection{3D Generation} 

Early methods on image-to-3D mainly employ Generative Adversarial Networks (GANs) to produce 3D representations, like point cloud~\cite{li2018point}, signed distance function (SDF)~\cite{shi2022deep, zou2023triplane, chan2022efficient} or neural radiance field (NeRF)~\cite{gu2021stylenerf, niemeyer2021giraffe}. By incorporating 2D GAN architectures \cite{brock2018large, creswell2018generative, kang2023scaling, karras2017progressive, karras2021alias, karras2019style, karras2020analyzing, zhu2023exploring}, 3D GANs have made remarkable advancements to generate 3D-aware content in a feed-forward manner. As a representative attempt, EG3D~\cite{chan2022efficient} employs
GANs to generate samples in latent space and introduce a compact triplane representation for rendering novel views of 3D objects. 
Another research branch in 3D generation utilizes diffusion models (DMs) \cite{tewari2024diffusion, ho2020denoising, rombach2022high, song2020score, po2023state, ddim, ddpm, ldm}, which have emerged with unprecedented success. One approach involves training 3D DMs directly using 3D supervision \cite{gupta20233dgen, jun2023shap, nichol2022point, ntavelis2023autodecoding, shue20233d} or 2D supervision~\cite{anciukevivcius2023renderdiffusion, chen2023single, gu2023nerfdiff, karnewar2023holodiffusion, liu2023zero, shen2023gina}. However, these approaches suffer from 3D inconsistency and poor quality due to the lack of 3D priors. DreamFusion~\cite{poole2022dreamfusion} introduces the Score Distillation Sampling (SDS) loss and generates NeRF under the supervision of pretrained large-scale 2D DMs. Variants of DreamFusion~\cite{chen2023fantasia3d, chung2023luciddreamer, hertz2023delta, liang2023luciddreamer, lin2023magic3d, poole2022dreamfusion, shi2023mvdream, tang2023dreamgaussian, wang2023score, wang2024prolificdreamer} aim to reduce Janus (multi-faced) artifacts and improve geometry quality. However, the prolonged optimization process limits its practical application.

\subsection{Sparse-View Reconstruction}

Multi-view images provide stereo cues facilitating more accurate and detailed shape generation. Consequently, sparse view-based neural reconstruction methods~\cite{chen2021mvsnerf, jain2021putting, lin2023vision, long2022sparseneus, wang2021ibrnet, yu2021pixelnerf} achieve significant improvements in generalization to unseen scenes. However, challenges remain in capturing patterns within large-scale datasets. Although scaling up the model and datasets is beneficial, it also introduces new challenges, such as increased computational complexity and sub-optimal generalizability. To address these challenges, recent methods leverage large-scale 3D datasets and well-designed multi-view diffusion models (DMs) to generate multiple novel views from a single image~\cite{shi2023mvdream, wang2023imagedream, long2023wonder3d, liu2023syncdreamer, tang2023MVDiffusion, tang2024mvdiffusion++, liu2023zero123, shi2023zero123++, one2345}. Subsequent steps involve neural reconstruction or Gaussian reconstruction~\cite{kerbl20233d, tang2024lgm, xu2024grm}. However, these explicit reconstruction methods face challenges due to multi-view inconsistency and low image resolution. Inspired by~\cite{hong2023lrm, wang2023pf, zou2023triplane}, we employ the recently proposed transformed-based large reconstruction model for implicit 3D shape generation.


\subsection{Large Reconstruction Model}
The extensive 3D datasets \cite{deitke2023objaverse, deitke2024objaverse} have opened up new possibilities for training reconstruction models that can create 3D models from 2D images. LRM~\cite{hong2023lrm} first showcased the potential of using the transformer backbone to map image tokens to implicit 3D triplanes with multi-view supervision, resulting in highly generalizable models. Instant3D~\cite{li2023instant3d} achieves highly generalizable and high-quality single-image to 3D generation by extending LRM to a sparse-view reconstruction model and utilizing the prior in 2D diffusion models. While LGM~\cite{tang2024lgm} and GRM~\cite{xu2024grm} models have replaced the triplane NeRF representation with 3D Gaussians \cite{kerbl20233d} to improve rendering efficiency, Gaussians have limitations in explicit geometry modeling and high-quality surface extraction. Concurrent works $\text{MVD}^2$~\cite{zheng2024mvd2}, CRM~\cite{wang2024crm}, and InstantMesh~\cite{xu2024instantmesh} use the same transformer backbone and focus on optimizing mesh representation for high-quality geometry and texture modeling. In contrast, our work focuses on incorporating 3D geometry prior to the time-consuming transformer attention for efficient training. Moreover, our proposed M-LRM, featuring a novel 3D-aware transformer block, allows significant generalization in 3D shape modeling.


\begin{figure*}[ht!]
\begin{center}
\includegraphics[width=1\linewidth]{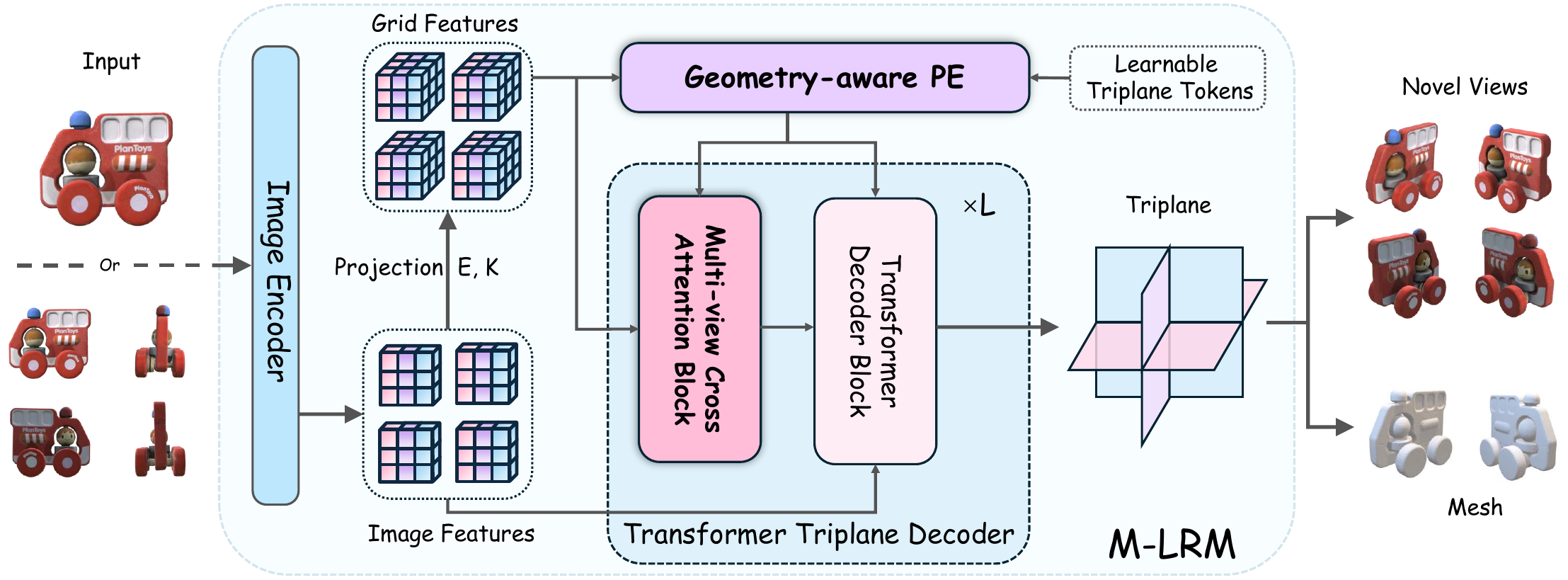}
\end{center}
\caption{\textbf{Overview of M-LRM}. M-LRM is a fully differentiable transformer-based framework, featuring an image encoder, geometry-aware position encoding and a multi-view cross attention block. Given multi-view images with corresponding camera extrinsic $E$ and intrinsic $K$, M-LRM incorporates the 2D and 3D features to conduct 3D-aware multi-view attention. Our transformer triplane decoder consists of $L$ layers, where each layer is composed of a multi-view cross attention Block and a vanilla transformer block. The proposed geometry-aware position encoding allows more detailed and realistic 3D generation.}
\label{fig:pip}
\vspace{-10pt}
\end{figure*}

\section{M-LRM}

Fig.~\ref{fig:pip} illustrates the overall framework of M-LRM. Unlike LRM which treats single-image-to-3D as a pure image translation problem, our method fully leverages the geometry prior to link 2D image and 3D triplane, achieving faster convergence and being able to generate results of better quality. Specifically, we achieve such advances in two steps. (1) Given multiple views of an object, we first extract the image features with an image encoder and propose \textit{Geometry-aware Positional Embeddings} to initialize the learnable triplane tokens. (2) With the learnable triplane tokens, we further adopt \textit{Geometry-aware Cross Attention} to achieve the image feature injection in a geometry-aware manner. Finally, we utilize the NeRF~\cite{mildenhall2021nerf} renderer to generate 3D contents with the generated triplane token. 

\subsection{Geometry-aware Positional Embeddings}
\label{Geometry-aware Positional Embeddings}

In LRM~\cite{hong2023lrm} and its variants~\cite{wang2023pflrm,li2023instant3d}, the triplane tokens are randomly initialized without any priors and directly optimized via back propagation. For convenience, we denote the randomly initialized triplane tokens as $\mathbf{T}_{r}$.
In this section, we propose Geometry-aware Positional Embedding (GaPE) to better utilize image prior instead of the randomly initialized triplane tokens.

\noindent \textbf{Geometry Volume Construction.} 
Inspired by the previous multi-view reconstruction method~\cite{long2022sparseneus}, we aim to directly learn a geometry volume from the input images which encodes local geometric information based on the hints of multi-view consistency of the images.
To cope with this, we first construct a grid volume whose vertices $\mathbf{X}=\{\mathbf{x}^j\}$ are uniformly sampled in the region of interest to be reconstructed, where $\mathbf{x}^j$ denotes the $j$-th vertex of the spatial grid volume. 

Considering the $i$-th view among the $N$ input views, with its intrinsic $K_i \in \mathcal{R}^{3\times 3}$ and extrinsic $E_i \in \mathcal{R}^{3\times 4}$, we back project the vertices to the feature map of the $i$-th view $F_i$,
and derive the corresponding feature grid ${G}_{i}=\{F_i(\pi_i(\mathbf{x}^j)) \}$ for the $i$-th view, where $\pi_i$ indicates the projection that projecting a 3D vertex onto the $i$-th image plane. The projection $\pi_i$ is defined as
\begin{equation}
\label{eq:projection}  
\Tilde{\mathbf{u}}^{j}_{i}  =\pi_i\left(\mathbf{x}^{j}\right) 
 =K_i E_i \Tilde{\mathbf{x}}^{j},
\end{equation}
where $\mathbf{x}^j \in \mathcal{R}^{3 \times 1}$ is the 3D coordinate of the grid vertex and $\Tilde{\mathbf{x}}^{j} \in \mathcal{R}^{3 \times 1}$ is the homogeneous coordinate of $\mathbf{x}^j$. 
$\mathbf{u}^{j}_{i} \in \mathcal{R}^{2 \times 1}$ is the 2D coordinate of 2D image space and 
$\Tilde{\mathbf{u}}^{j}_{i} \in \mathcal{R}^{3 \times 1}$ is the homogeneous coordinate of $\mathbf{u}_i^j$.
With the obtained 2D coordinates, we could sample corresponding features from the 2D feature map $F_i$.
After back-projecting all the grid vertices, we obtain $N$ feature grids $\{G_i\}_{i = 1, \cdot\cdot\cdot, N}$, which are fused together into one unified feature volume $\mathbf{G}$ with a 3D convolution network.

\noindent \textbf{Geometric Triplane Token.}
The feature volume $\mathbf{G}$ contains the enriched prior produced by the image encoder. To introduce such prior to the triplane tokens, for each axis, we collect all the features of the volume along the axis, where the features of different vertices are concatenated, and obtain three feature maps $\mathbf{g}^{xy}, \mathbf{g}^{yz}, \mathbf{g}^{xz}$. The feature maps are further fed into three unique 2D convolutional networks to produce the positional embedding $\mathbf{T}_{g}^{xy}, \mathbf{T}_{g}^{yz}, \mathbf{T}_{g}^{xz}$. 
Finally, fusing the positional embedding and randomly initialized triplane tokens $\mathbf{T}_{r}$ together yields the final geometry-aware triplane tokens as
\begin{equation}
    \mathbf{T}_0 = \mathbf{T}_{g} + \mathbf{T}_{r}.
\end{equation}
Since the tokens explicitly incorporate the geometric priors of the geometry volume, the tokens are geometry-aware and lead to faster convergence.


\begin{figure*}[ht]
\begin{center}
\includegraphics[width=0.8\linewidth]{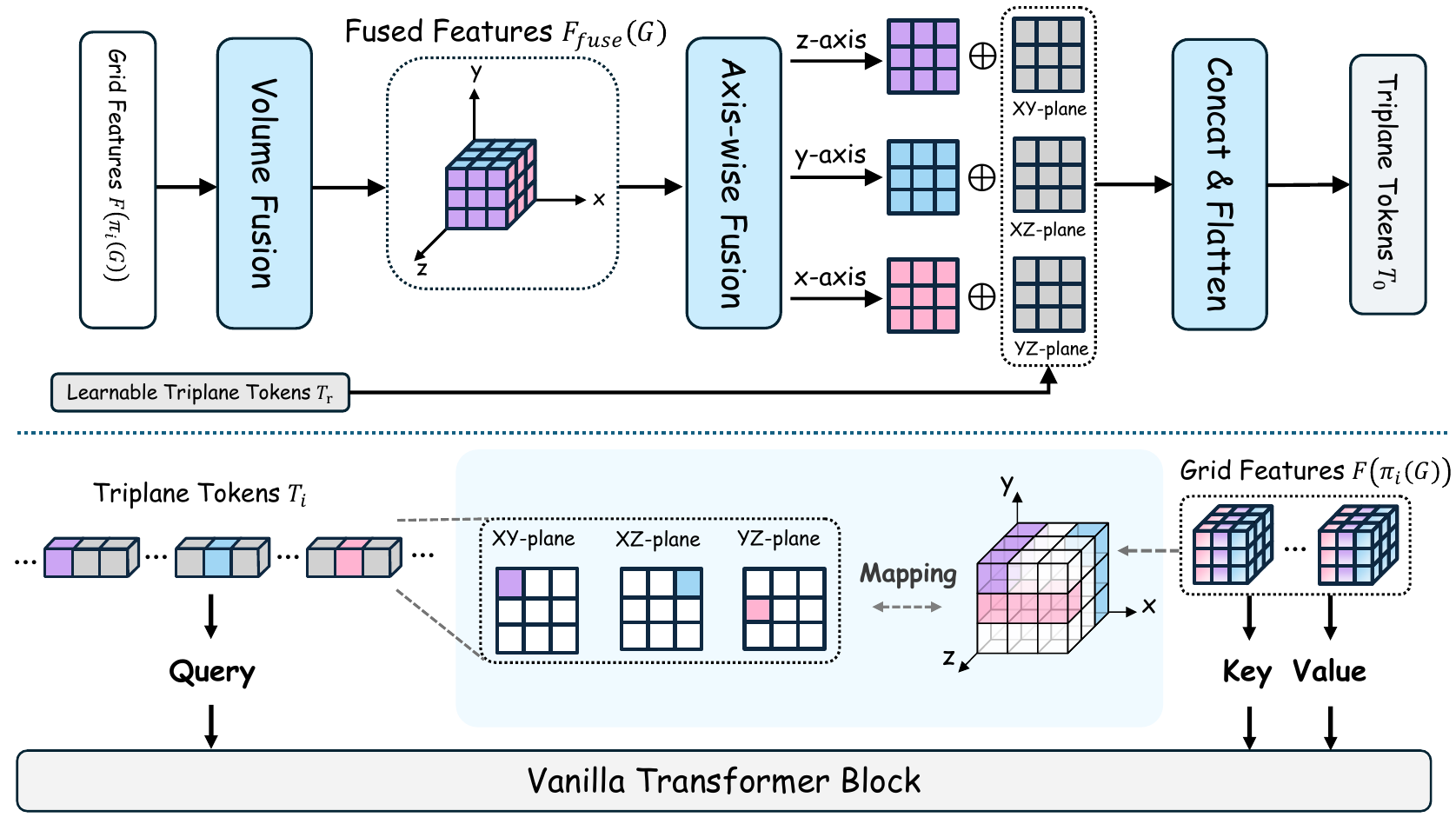}
\caption{Pipeline of Geometry-aware Positional Embedding(GaPE) and Geometry-aware Cross Attention(GCA). GaPE~(top) fuse the enriched priors of input views in the grid features and randomly initialized triplane tokens to make the triplane tokens geometrically aware. GCA~(bottom) use triplane tokens as query and find the spatially corresponding tokens in each feature grid as key and value to incorporates 3D prior and interrelations of the multiple input views.}
\end{center}
\vspace{-0.5em}
\label{fig:figure2}
\vspace{-10pt}
\end{figure*}

\subsection{Geometry-aware Cross Attention}

\label{Multi-view Consistent Cross Attention}

\begin{figure}[t]
\begin{center}
\includegraphics[width=0.8\linewidth]{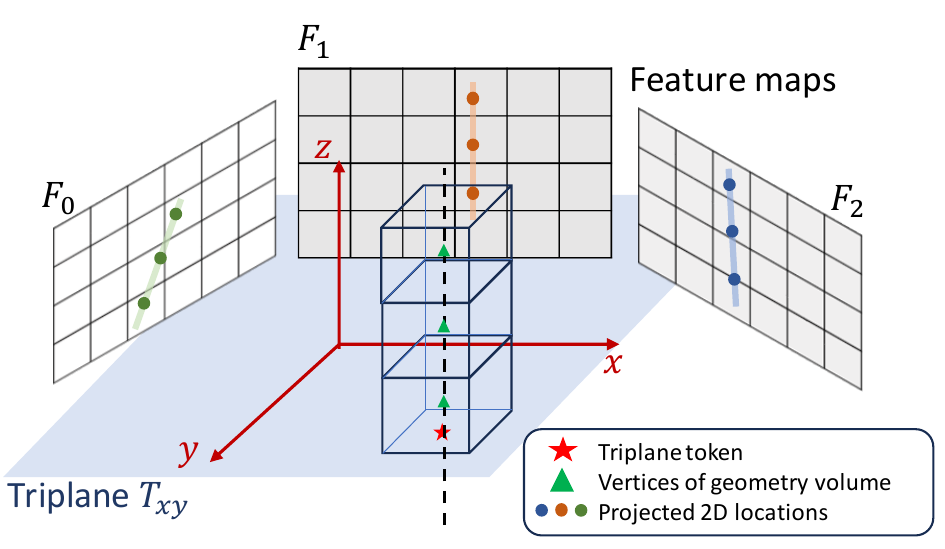}
\end{center}
\caption{\textbf{The illustration of Geometry-aware Cross Attention.} The vertices of geometry volume are projected back onto the 2D images. The features corresponding to the projected 2D locations are sampled as key and value in GCA.}
\label{fig:attention}
\vspace{-10pt}
\end{figure}

In addition to the geometry-aware triplane tokens, we introduce a novel cross-attention mechanism, named \textit{Geometry-aware Cross Attention} (GCA), which incorporates explicit 3D-aware priors for enhanced information fusion between image features and triplane tokens. Specifically, unlike the vanilla attention used in previous LRM works~\cite{hong2023lrm,wang2023pflrm,li2023instant3d} that accesses global features from all views, our method constructs an multi-view geometry-aware attention mechanism. This mechanism ensures that each triplane token correlates with only a limited number of image feature tokens, improving computational efficiency and relevance.



Intuitively, each token in a triplane corresponds to a ray in the 3D space passing through the token, which can be projected into each view (as illustrated in Fig.~\ref{fig:attention}). Consequently, when performing cross-attention between each triplane token and the image features, we only need to correlate the triplane token with the feature tokens intersecting each projected ray.
In prior works, dense cross-attention operates on $N \times h \times w$ feature tokens, where $N$ is the number of views, and $h$ and $w$ are the height and width of the feature map. In contrast, our geometry-aware attention operates on only $N \times V$ feature tokens, where $V$ is the size of the geometry volume. This significantly reduces the computational complexity and focuses attention on more relevant features.

The key to geometry-aware attention is extracting the feature tokens that intersect with each projected ray. By retaining the grid volume as mentioned, we can leverage Eq.~\ref{eq:projection} to project the relevant volume vertices into each view and obtain the corresponding 2D locations. Using these projected 2D locations, we can then access the related feature tokens for cross-attention. This approach ensures that the attention mechanism focuses on the most pertinent features, enhancing efficiency and relevance in the information fusion process.


Specifically, using a triplane token $t$ as query, we gather all vertices of the grid volume that is corresponding to $t$, then project these vertices to every 2D images as illustrated in Figure \ref{fig:attention}. Features are sampled according to the projected 2D locations and are used as key and value in our GCA module.





To preserve global information interaction among different triplane tokens, we only replace half of the original transformer blocks with GCA.



\begin{table*}[ht]
\centering
\caption{Comparison of multi-view reconstruction with state-of-the-art image-to-3D counterparts on the Google Scanned Objects (GSO) dataset.
$\uparrow$ means the higher the better, and $\downarrow$ indicates the lower the better. 
}
\resizebox{\textwidth}{!}{%
\footnotesize
\begin{tabular}{ccccccccc}
\toprule
Method                 & Data & Train Res. & PSNR$\uparrow$  & SSIM$\uparrow$  & LPIPS$\downarrow$ & CD$\downarrow$ & FS (t=0.001)$\uparrow$ & FS (t=0.01)$\uparrow$ \\ \midrule
Instant3D   &  Objaverse-LVIS    & 256        & 22.18 & 0.958 & 0.142 & 0.0055 & 0.7012 & 0.9167 \\
LGM                    &  Objaverse    & 512        & 21.74 & 0.967 & 0.110 & 0.0034 & 0.7108 & 0.9273 \\
Ours (Tiny)             &  Objaverse-LVIS   & 256        & 24.35 & 0.961 & 0.142 & 0.0021 & 0.7327 & 0.9491\\
Ours (Base)            &  Objaverse   & 512        & \textbf{25.23} & \textbf{0.977} & \textbf{0.096} & \textbf{0.0011}& \textbf{0.8725} & \textbf{0.9695} \\ \bottomrule
\end{tabular}%
}
\label{results_mv_recon}
\end{table*}

\subsection{Training Objectives}
\label{sec:train_obj}

After obtaining the decoded triplane, we adopt NeRF to render colors $\hat{I}$ and mask images $\hat{m}$. 
The predicted novel views are supervised via differentiable rendering by other sampled views. We adopt MSE loss, BCE mask loss and LPIPS loss \cite{zhang2018unreasonable}. The training loss can be formulated as


\begin{equation}
\begin{gathered}
\mathcal{L}_{mse}=\sum\limits_{i=1}^N\Big|\Big|\hat{I}_i-I_i\Big|\Big|_2 \\
\mathcal{L}_{lpips}=\sum\limits_{i=1}^N\mathcal{L}_{lpips}(\hat{I}_i, I_i) \\
\mathcal{L}_{mask}=\sum\limits_{i=1}^N m\log \hat{m}+(1-m)\log(1-\hat{m}) \\
\mathcal{L}_{NeRF}=\lambda_1\mathcal{L}_{mse}+\lambda_2\mathcal{L}_{lpips}+\lambda_3\mathcal{L}_{mask}
\end{gathered}
\end{equation}
where $I_i$ and $m_i$ denote the i-th rendered ground truth images and $\lambda_i~$(i=1,2,3) are the corresponding weight of MSE, LPIPS and mask loss. The coefficients for different loss components are $\lambda_{1} = 1.0$, $\lambda_{2} = 0.1$ and $\lambda_{3} = 1.0$.

\begin{figure*}[ht]
\centering
\includegraphics[width=0.8\linewidth]{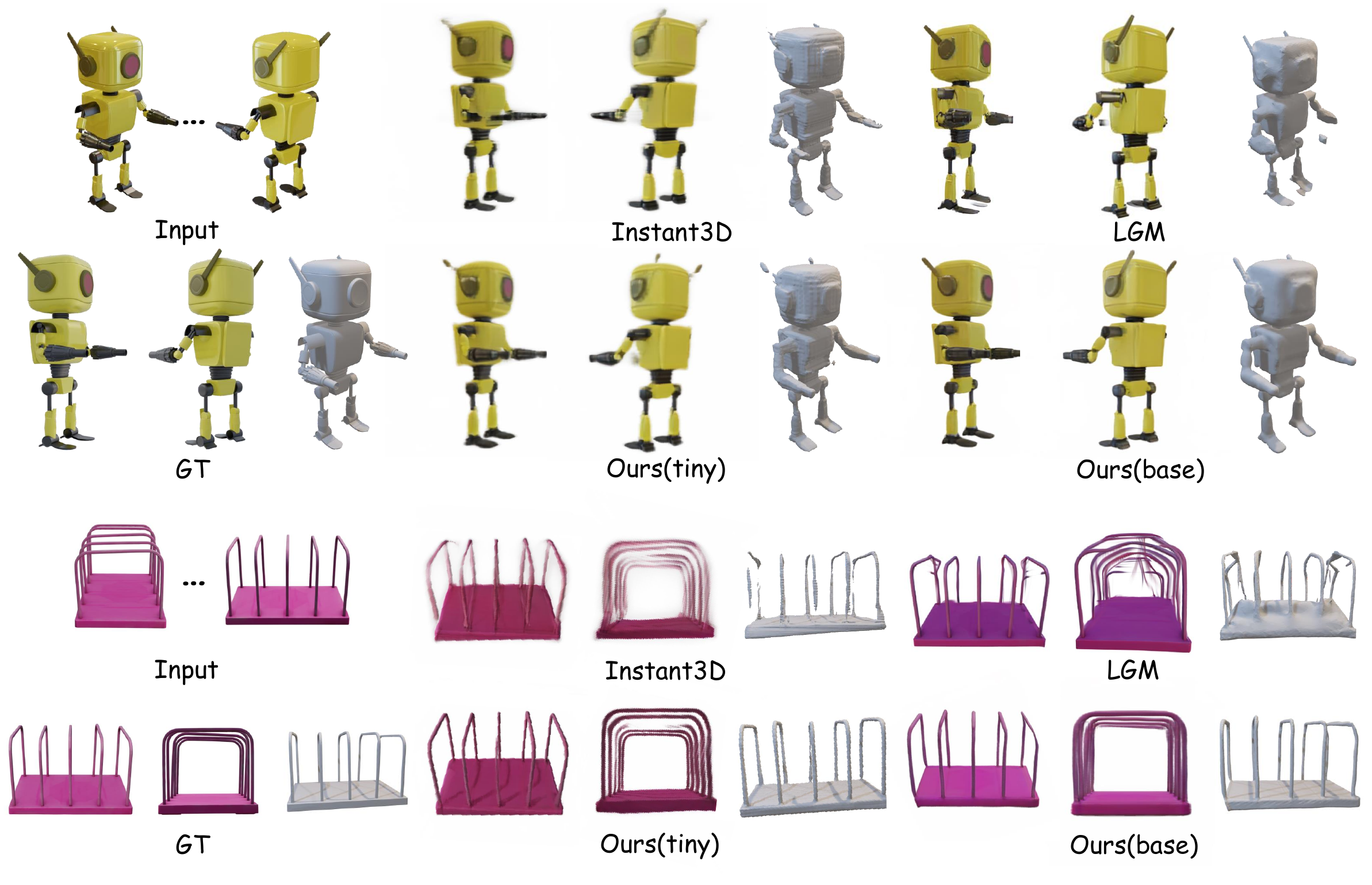}
\vspace{-0.5em}
\caption{\textbf{Qualitative result of sparse-view reconstruction.} We compare our generated results with two baseline methods. Given the same input conditional views, our tiny model can generate results of higher fidelity and geometries that are more reasonable. Our base variant uses two additional views as input (the rightmost two images of the input views).}
\label{fig:mv_recon}
\end{figure*}

\section{Implementation details} 

\subsection{Data Preparation}

\noindent \textbf{Datasets.}
We use the Objaverse\cite{deitke2023objaverse} dataset to train our base model, where 190K objects are filtered out among the 800K 3D models in the original dataset. We remove the objects without texture and the objects with tags indicating low quality. 

For our reproduced Instant3D, tiny model and all the models in our ablation study, we opt to use Objaverse-LVIS, a subset of Objaverse, due to the limitation of computational cost. 31K objects in Objaverse-LVIS are used for training.

Commonly used dataset Google Scanned Objects(GSO)\cite{downs2022google} is adopted as our evaluation set following previous work\cite{li2023instant3d}. All 1030 objects from GSO are used for evaluation.

\begin{figure*}[ht!]
\centering
\includegraphics[width=0.95\linewidth]{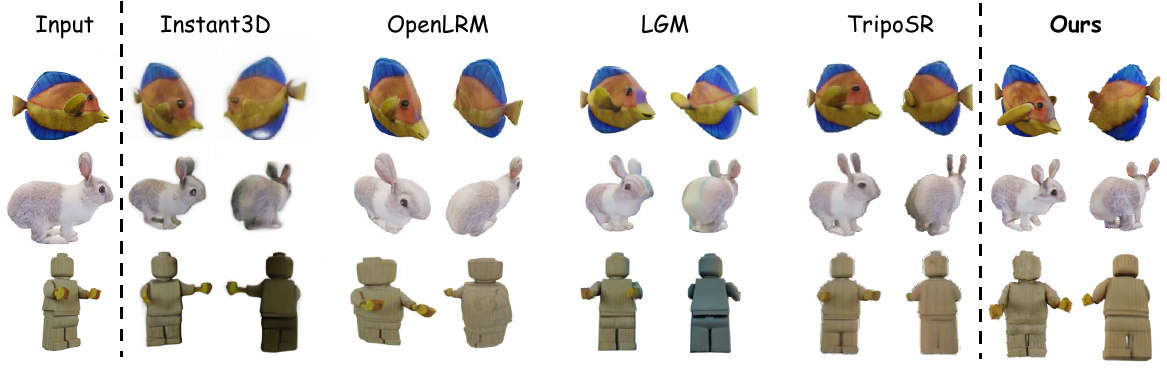}
\vspace{-0.5em}
\caption{\textbf{Visualization result of single view to 3D generation.} We compare our method with Instant3D \cite{li2023instant3d}, LGM \cite{tang2024lgm} and TripoSR \cite{TripoSR2024}. The results of Instant3D suffer from blurry texture and severe geometry failure, while LGM \cite{tang2024lgm} produces results with unreasonable color at the unseen region of the object. TripoSR can also generate visual results of high quality, but there remain many geometry artifacts. Our overall result is of better visual quality and geometry quality.}
\label{fig:single_view_results}
\vspace{-0.5em}
\end{figure*}

\noindent \textbf{Rendering Strategy.}
\label{rendering strategy}
We use Blender to render the images. The 3D models are scaled to fit into the cubic bounding box, whose length is 1 and center is  $(0, 0, 0)$. We set the field-of-view to $30^\circ$ and the camera distance is calculated accordingly to ensure the object is well-fitted at the center of the view. For Objaverse-LVIS and GSO, we render 24 views for each object. The first 8 views are sampled uniformly around the object with $0^\circ$ elevation, while the second 8 views are sampled uniformly with $20^\circ$ elevation. The last 8 views are sampled randomly on the surface of the sphere. For our Objaverse dataset, we render 32 views for each object. In addition to the 24 views mentioned before, we render 8 extra views that are sampled uniformly around the object with elevation $-20^\circ$.All cameras face towards $(0, 0, 0)$. Uniform lighting is adopted in all render processes. All images are rendered at the resolution of $512\times 512$.

\subsection{Training Scheme}

It takes about 3 days to train our base M-LRM on 32 NVIDIA H800 GPU (80GB) for 20 epochs. The batch size for each GPU was set to 1, and gradient accumulation was employed for 8 steps, resulting in a total batch size of 512. At each iteration, we randomly sample 10 views among the total 32 images, where 6 of them are used as input conditional images, and the other 4 for supervision. AdamW optimizer is used with $\beta_1, \beta_2$ set to $0.9, 0.95$ respectively, and weight decay is set to 0.05. We increase the learning rate from 4e-10 to 4e-4 for the first epoch. After warmup, we adopt cosine annealing strategy for the learning rate and the learning rate decay drops to 0 at the end of training. The coefficients for different loss components are $\lambda_{1} = 1.0$, $\lambda_{2} = 0.1$ and $\lambda_{3} = 1.0$. Following LGM \cite{tang2024lgm}, we adopt camera augmentation and image grid distortion as data augmentation. In each iteration, we randomly add noise to the input conditional camera poses(elevation, azimuth and camera distance) and apply grid distortion. The Gaussian noises are ranged in $[-10, 10]$ for elevation and azimuth and $[-0.1, 0.1]$ for camera distance. 50\% of the iterations are affected by the data augmentation.

For other models that are trained on Objaverse-LVIS, we set the batch size to 8 for each GPU, leading to a batch size of 64 in total with 8 H800 GPUs. We train these models for 100 epochs. The other hyper-parameters are the same.

\subsection{Network Architecture}

To implement our base M-LRM, we utilize DINOv2-base with a patch size of 14 as our image encoder. The resolution for the 6 input images is $504\times 504$. The transformer triplane decoder consists of 6 layers with a hidden dimension of 768. We employ multi-head attention with 8 heads, and each head has a hidden dimension of 64. The learnable triplane tokens are initialized with a resolution of $64\times64$ and 768 channels. To upsample the triplane, we use a transpose convolution layer with a kernel size of 2 and a stride of 2, resulting in a resolution of 128x128 and 40 channels. The concatenated triplane features are then decoded using a 4-layer MLP into density and RGB values. We sample 192 points per ray. The model is supervised under the resolution of $512\times 512$. At each iteration, a randomly cropped window of size $128\times 128$ is sampled for supervision for saving GPU memory.

We construct the geometry volume with grid size $V=64$, yielding a resolution of $64^3$. After projecting the image features into spatial volumes, we aggregate them using a 5-layer 3D convolution network with channel sizes of [64, 128, 128, 128, 64] into feature volume $\mathbf{G}$. We derive our positional embedding $\mathbf{T}_g$ from the volume $\mathbf{G}$ using three 2D convolution layers with a kernel size of 3 and a stride of 1.

For our reproduced Instant3D, tiny model and all the models in our ablation study, we use 4 views as input. All images are resized to $256\times 256$ for supervision. The transformer triplane decoder of our tiny model consists of 5 layers instead of 6. The decoder of our reproduced Instant3D has 10 layers of transformer blocks. Camera augmentation and grid distortion augmentation are disabled. All other settings are the same as our base model.





\section{Experiments}

\begin{figure*}[]
\includegraphics[width=\linewidth]{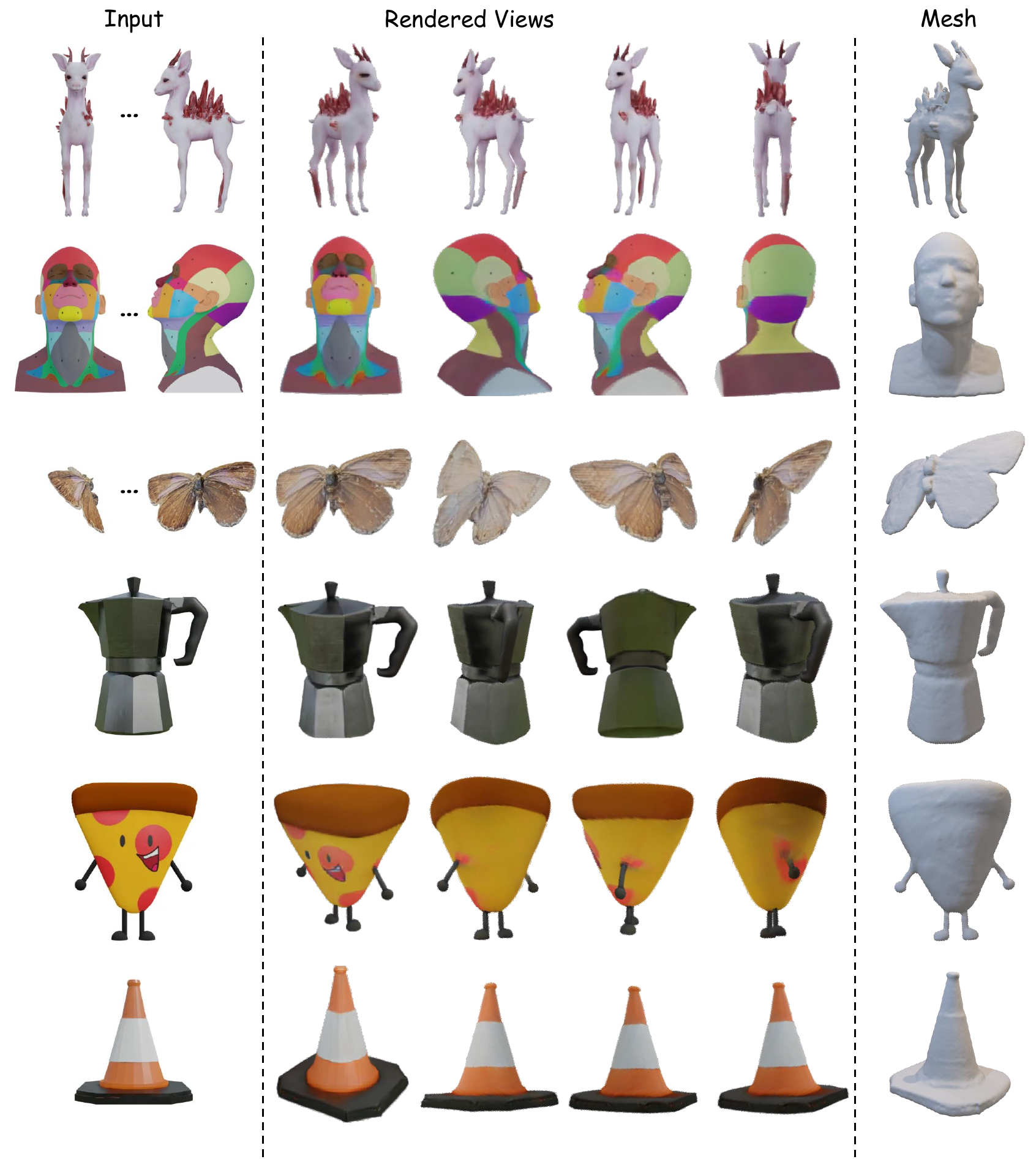}
\caption{\textbf{More visualization results of single-view generation and multi-view reconstruction.} Given multiple images, our M-LRM is capable of reconstructing the object. The high fidelity novel views and high quality generated mesh demonstrates the superiority of our proposed method.Given a single image, we first synthesize the multi-view images with multi-view generation model, than feed the images into our M-LRM. Our proposed method facilitates generating novel views and meshes.}
\label{fig:mv_sv_result}
\end{figure*}




\subsection{Multi-view Reconstruction}
\label{MV_recon}

\noindent \textbf{Evaluation Setting.}
We compare our methods with Instant3D~\cite{li2023instant3d} and LGM~\cite{tang2024lgm} on Google Scanned Objects (GSO) \cite{downs2022google}. Since the code and the model of Instant3D \cite{li2023instant3d} are not available, we reproduce the model according to their paper.
To fairly compare with Instant3D, we also train a tiny variant of our model that uses the same data as our reproduced Instant3D. As for the comparison with LGM, we use the original implementation and default hyper-parameter setting of LGM for the sparse view reconstruction task. Following existing work, we report PSNR, SSIM, and LPIPS to evaluate the visual quality of reconstruction and Chamfer distance (CD), F-Score(FS) for measuring the quality of the reconstructed mesh. We uniformly sample 16K points for measuring CD. We use two different thresholds $t$ to calculate FS for evaluation. All images are rendered at the resolution of $512 \times 512$ for evaluation.

For all 1030 objects in GSO, we render 24 views following the same setting described in the rendering strategy(Sec. \ref{rendering strategy}). Four views with elevation $0^\circ$ and azimuth $[0, 90^\circ, 180^\circ, 270^\circ]$ are input for all the 4-view models. For our base model, two extra views with elevation $20^\circ$ and azimuth $[45^\circ, 225^\circ]$ are used as input conditional images. The rest views are utilized to calculate the average PSNR, SSIM, and LPIPS.

\noindent \textbf{Results.}
As shown in Tab. \ref{results_mv_recon}, our proposed M-LRM outperforms all baselines by a large margin. Even though our base version uses the most views compared to previous methods, our tiny version model still demonstrates the superiority of our proposed method. Although our tiny model is trained with much less data and lower resolution than LGM, it can still generate images and meshes of higher quality. However, the lack of data makes the model fail to generalize to the evaluation dataset, leading to higher LPIPS. When our M-LRM is equipped with more input views and more training data, it can produce 3D objects that are both visually and geometrically plausible, as demonstrated by the results of our base model.

Fig. \ref{fig:mv_recon} shows some visualization results of sparse-view reconstruction on the GSO dataset. Our method can not only generate novel views of better visual quality with no floaters but also is able to extract meshes that are more plausible. Instant3D and LGM generate blurry novel views and the meshes they generate suffer from severe geometric failure. This indicates that LRM's learning ability is limited without explicit 3D priors. With our proposed method, our tiny model successfully obtains the geometric reconstruction ability with limited data. When we extend the model to more input views and more training data, our base model can reconstruct high-quality objects.

\subsection{Single Image to 3D Generation}
\label{SV_gen}

\begin{table}[]
\centering
\vspace{-10pt}
\caption{Comparison of single-view generation with different single-view generation methods on the Google Scanned Objects(GSO) dataset.}
\resizebox{\columnwidth}{!}{%
\begin{tabular}{cccc}
\hline
Method         & CD$\downarrow$              & FS(t=0.001)$\uparrow$     & FS(t=0.01)$\uparrow$      \\ \hline
Instant3D      & 0.0093          & 0.4597          & 0.7948          \\
TripoSR        & 0.0121          & 0.4548          & 0.6879          \\
OpenLRM        & 0.0060          & 0.5116          & 0.8353          \\
LGM            & 0.0086          & 0.4650          & 0.7841          \\
LGM(Zero123++) & 0.0062          & 0.4558          & 0.8315          \\
Ours(tiny)     & 0.0055          & 0.5614          & 0.8465          \\
Ours(base)     & \textbf{0.0045} & \textbf{0.6148} & \textbf{0.8744} \\ \hline
\end{tabular}%
}
\label{results_sv_gen}
\vspace{-20pt}
\end{table}

\noindent \textbf{Evaluation Setting.}
For single image to 3D generation, we adopt Zero123++ \cite{shi2023zero123++} to generate multiple views as conditional images. For our tiny model and our reproduced Instant3D that takes 4 views as input, we select the first, third, fifth, and sixth images generated by Zero123++ following GRM\cite{xu2024grm}. We compare our methods with Instant3D~\cite{li2023instant3d}, TripoSR~\cite{TripoSR2024}, OpenLRM~\cite{openlrm}, and LGM~\cite{tang2024lgm} qualitatively(Figure. \ref{fig:single_view_results}) and quantitatively(Table. \ref{results_sv_gen}). For LGM that uses external multi-view generation models to obtain multi-view images for reconstruction, we additionally adopt the same Zero123++ model as ours to test LGM for fair quantitative comparison. 

We adopt GSO dataset for quantitative comparison. For each object in GSO, we collect the front view with elevation $0^\circ$ and azimuth $0^\circ$ as input to generate the multi-view images and mesh. After generating multi-view images with Zero123++ and reconstructing with our M-LRM, we extract the mesh with with density field via marching cube. The generated mesh is aligned with the GT mesh by first conducting ICP algorithm and then resizing the generated mesh to the same scale as GT. We report Chamfer distance(CD) and F-Score(FS). 16K points are sampled uniformly to calculate these metrics. Two different thresholds are adopted following the evaluation setting of our multi-view reconstruction task.

\noindent \textbf{Results.}
We compare the qualitative results of our generated results with those of baseline methods. As shown in Fig. \ref{fig:single_view_results}, we use three examples to explain our qualitative experiment. The generation outcomes of Instant3D exhibit deficiencies in texture clarity (cases 1 and 3). OpenLRM\cite{openlrm} suffers from severe geometric faults(cases 1, 2, 3), because it lacks the modeling of the invisible parts of the objects. Similarly, the results produced by LGM \cite{tang2024lgm} display unreasonable texture color in the unobserved regions of the object (cases 2, 3). Besides, the geometry produced by LGM contains some geometry artifacts(case 1). Although TripoSR\cite{TripoSR2024} is capable of generating visually appealing results, it still exhibits numerous imperfections in geometric accuracy (case 3) and inconsistency with input image(cases 1, 2) due to the lack of explicit 3D spatial modeling. In contrast, as our proposed method is geometry-aware, it yields superior visual quality and improved geometric fidelity.

The results of quantitative comparison are shown in Table. \ref{results_sv_gen}. Our proposed M-LRM outperforms all previous method. Since LGM also uses an external multi-view generation framework before generating the mesh, we replace the framework used by LGM with Zero123++ for fair comparison. Although it slightly improves the generated mesh of LGM due to better consistency of the generated views, our M-LRM still generates the best results. Even with limited training data, our tiny model shows fast convergence speed and strong abilities in 3D reasoning, thus generating meshes of high quality. With sufficient data, our base model demonstrates strong generation ability.

\subsection{Effectiveness of Model Components}
\label{ablation}

\begin{table}[]
\centering
\vspace{-10pt}
\caption{The ablation study of each component.}
\resizebox{\columnwidth}{!}{%
\tiny
\begin{tabular}{ccccccccc}
\hline
Model   & GCA       & GaPE      & w/ Tr. Block & High Res. & Large & PSNR$\uparrow$                       & SSIM$\uparrow$                       & LPIPS$\downarrow$                      \\ \hline
\textbf{(1)}&\ding{56} & \ding{56} & \ding{56}    & \ding{56} & \ding{56} & 24.501                     & 0.9746                     & 0.1364                     \\
\textbf{(2)}&\ding{52} & \ding{56} & \ding{56}    & \ding{56} & \ding{56} & 24.923 & 0.9750 & 0.1322 \\
\textbf{(3)}&\ding{56} & \ding{52} & \ding{56}    & \ding{56} & \ding{56} & 25.116 & 0.9765 & 0.1269 \\
\textbf{(4)}&\ding{52} & \ding{52} & \ding{56}    & \ding{56} & \ding{56} & 25.184 & 0.9770 & 0.1233 \\
\textbf{(5)}&\ding{52} & \ding{52} & \ding{52}    & \ding{56} & \ding{56} & 25.254                     & 0.9772                     & 0.1246                     \\
\textbf{(6)}&\ding{56} & \ding{56} & \ding{56}    & \ding{52} & \ding{56} & 26.219 & 0.9756 & 0.1201 \\
\textbf{(7)}&\ding{56} & \ding{56} & \ding{56}    & \ding{52} & \ding{52} & 26.588 & 0.9772 & 0.1136 \\

\textbf{(8)}&\ding{52} & \ding{56} & \ding{52}    & \ding{52} & \ding{56} & 26.966                     & 0.9781                     & 0.1067                     \\
\textbf{(9)}&\ding{52} & \ding{52} & \ding{52}    & \ding{52} & \ding{56} & \textbf{27.581}            & \textbf{0.9803}            & \textbf{0.0986}            \\ \hline
\end{tabular}%
}
\label{ablation_result}
\vspace{-10pt}
\end{table}

We conduct ablation studies to demonstrate the effectiveness of our model components and architectural designs. We train these model variants with only Objaverse-LVIS subset. These models are evaluated on our randomly picked test set from Objaverse-LVIS. We ensure that all data in this test set do not appear in the training set. Our ablation study follows the evaluation setting of multi-view reconstruction, but uses the test split in Objaverse-LVIS. 3K objects are randomly chosen from Objaverse-LVIS as our test split. 

As shown in Tab.~\ref{ablation_result}, GCA means using our proposed Multi-view Cross Attention module, GaPE means using Geometry-aware Positional Encoding, w/ Tr. Block means adding a transformer decoder block after our MCA block, and High Res. means using 64 as the resolution of the initial triplane and the geometry volume, as opposed to the original 32. When High Res. is adopted, we sample 192 points per ray in volumetric rendering instead of 96 originally. The large in the table indicates that we extend the original Instant3D model to have the equivalent number of parameters to ours(tiny). This is to ensure fair comparison and prove that the performance gain comes from our proposed components.

The results of the model (1), (2), (3) and (4) demonstrate the effectiveness of our proposed GCA and GaPE. As shown by the results of models (4) and (5), adding a transformer decoder block after GCA block improves the model's performance, indicating that preserving global attention is helpful. We also show that using a high-resolution setting enhances the model's capability significantly, as shown by models (1) and (6). We also prove that the performance gain of our methods does not come from the extra parameters introduced, as demonstrated by models (7) and (9). Models (6), (8), and (9) indicate that even under high-resolution setting, our proposed methods are still effective. 

\subsection{Robustness}
\begin{table*}[ht]
\centering
\caption{Experiments with inaccurate camera parameters as input. We test LGM and our base model with inaccurate camera intrinsics and poses. $s$ is the pre-defined factor to control the scale of the noise added.}
\resizebox{\textwidth}{!}{%
\begin{tabular}{ccccc}
\hline
Method     & PNSR/SSIM/LPIPS(s=0)         & PNSR/SSIM/LPIPS(s=2)         & PNSR/SSIM/LPIPS(s=5)         & PNSR/SSIM/LPIPS(s=10)        \\ \hline
LGM        & 21.74/0.9668/0.1104          & 21.41/0.9657/0.1137          & 20.34/0.9607/0.1288          & 18.78/0.9522/0.1458          \\
Ours(base) & \textbf{25.23/0.9773/0.0961} & \textbf{23.68/0.9746/0.0996} & \textbf{21.22/0.9672/0.1171} & \textbf{18.92/0.9562/0.1412} \\ \hline
\end{tabular}%
\label{tab:robust}
}
\end{table*}

Since the GCA and GaPE modules we introduced in this paper highly rely on the condition camera intrinsic and poses, we conduct extra experiments to test our model's robustness against inaccurate camera parameters and out-of-domain camera distributions. 

Following our multi-view generation evaluation setting, we add extra random Gaussian noise to the elevation, azimuth, camera distance, and focal length of the condition cameras to test our base model and LGM. The results show that to some extent, our model is robust to the inaccurate camera information. However, we point out that the robustness of our model is still limited. Robust reconstruction with inaccurate camera parameters is still an open challenge.

As shown in Table \ref{tab:robust}, $s$ is the noise strength. For elevation and azimuth, the noise is ranged in $[-s, s]$. For camera distance, the noise is within $[-0.01s, 0.01s]$. And for focal length, the noise in ranged in $[-10s, 10s]$. Both LGM and our M-LRM suffer from performance drop given inaccurate camera intrinsics and poses as input. However, our method still performs better than LGM in all noise strength settings.

The results show that, to some extent, our model is robust to inaccurate camera parameters and can generalize to different camera distributions.

\begin{table}[h]
\centering
\caption{Experiment results with different camera distribution. We use cameras of different FOV and camera distance to test our base model. The results show that our model can generalize to different camera distributions although trained with fixed FOV and camera distance}.
\resizebox{\columnwidth}{!}{%
\begin{tabular}{ccccccc}
\hline
Camera Distribution & PSNR           & SSIM           & LPIPS          & CD              & FS(t=0.001)     & FS(t=0.01)      \\ \hline
Default             & \textbf{25.23} & \textbf{0.977} & \textbf{0.096} & \textbf{0.0011} & \textbf{0.8725} & \textbf{0.9695} \\
Random              & 24.86          & 0.976          & 0.101          & 0.0018          & 0.7405          & 0.9556         \\ \hline
\end{tabular}%
\label{tab:generalize}
}
\end{table}

To test our base model's generalizability when the given camera's distribution is different from the one used for training, i.e. different FOV and camera distances, we re-render the evaluation dataset under a different setting. For every object in the GSO dataset, we choose a random FOV between 15 and 45, then select an appropriate camera distance accordingly to make sure that the object is well-fitted into the screen. The camera distance is randomly scaled with a factor ranging in $[0.9, 1.2]$. Other settings are the same as our default render setting. The results show that although our base model is trained with fixed FOV and camera distance following Instant3D and LGM, our model shows strong generalizability against different camera distributions. The results are shown in Table \ref{tab:generalize}.

\subsection{Convergence Speed}

\begin{figure*}[t]
\includegraphics[width=\textwidth]{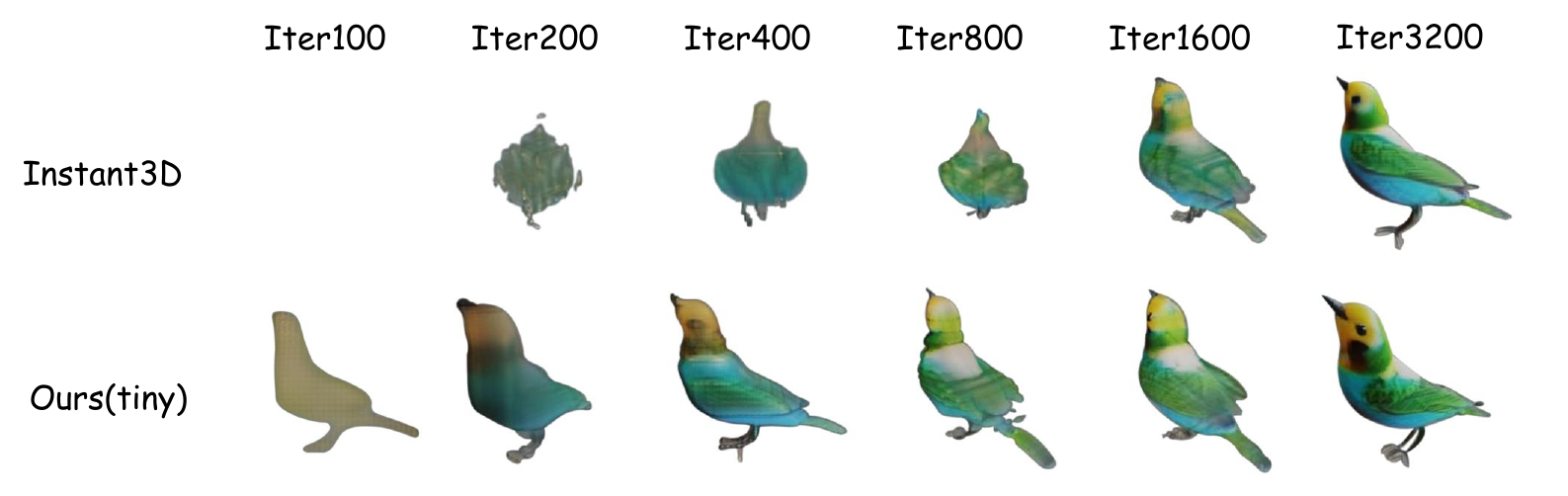}
\caption{\textbf{Visualization of intermediate results during training process.} Compared with existing LRM-based approaches, like Instant3D~\cite{li2023instant3d}, our proposed GCA and GaPE remarkably reduce the convergence time and improve the reconstruction quality by a large margin. At the early stage of the training, our model shows stronger fitting ability and converges faster than the baseline method Instant3D. When trained for sufficient time, our proposed GaPE and GCA enhance the model's reconstruction ability, yielding better generation results.}
\label{fig:convergence}
\end{figure*}

\begin{figure}[t]
\includegraphics[width=0.9\linewidth]{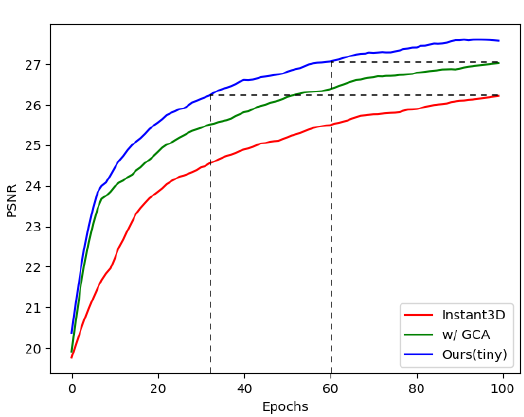}
\vspace{-10pt}
\caption{\textbf{PSNR curve of different models on the Objaverse-LVIS validation set.} Compared to the baseline method Instant3D, our proposed GCA and GaPE significantly facilitate the convergence of LRM model. The introduction of each module greatly speed up the convergence speed enhance the model's ability when trained for the same time.}
\label{fig:converge_curve}
\vspace{-20pt}
\end{figure}

To showcase our superior convergence speed, we visualize the intermediate results during the training process of our reproduced Instant3D and our tiny model, as they are trained under the same setting. As shown in Figure \ref{fig:convergence}, in both cases, our model converges significantly faster than Instant3D. At the very early stage of the training(400 iterations), our model starts to produce plausible and reasonable geometries, while Instant3D fails to generate satisfactory results until the later training stage. 

Our quantitative results in Table \ref{results_mv_recon} also showcase the fast convergence and learning ability of our M-LRM. With only Objaverse-LVIS subset, our tiny model can achieve better reconstruction results than LGM that use Objaverse dataset instead. 

As the PSNR curve on validation set presented in Figure \ref{fig:converge_curve}, our proposed GCA and GaPE significantly accelerate the convergence process. When GCA is implemented (green curve), our M-LRM demonstrates comparable generation capabilities to Instant3D (red curve) with only half the number of training iterations. Furthermore, incorporating both GCA and GaPE (blue curve) reduces the required epochs to one-third. When these models are trained for an equivalent number of epochs, our M-LRM significantly surpasses the performance of the other two models.

\subsection{Inference Speed}

\begin{table}[h]
\centering
\caption{Results of inference speed comparison. We compare the speed of 3D representation inference, mesh extraction and image rendering of different methods.}
\resizebox{\columnwidth}{!}{%
\begin{tabular}{cccc}
\hline
Method           & 3D Representation/s & Mesh Extraction/s & View Rendering/s \\ \hline
Instant3D        & 0.6              & 1.5            & 0.075         \\
Instant3D(large) & 0.8              & 1.5            & 0.075         \\
LGM              & 0.2              & ~300          & 0.01          \\
Ours(tiny)       & 0.8              & 1.5            & 0.075         \\
Ours(base)       & 1                & 1.5            & 0.08         \\ \hline
\end{tabular}%
\label{tab:inference}
\vspace{-30pt}
}
\end{table}

Although our proposed methods introduced more parameters and operations, which might increase the computational cost of our model, these additional costs are not the bottleneck of the model's inference performance. We compare the inference speed of our model and baseline methods in Table \ref{tab:inference}. Our M-LRM does inference more slowly than Instant3D due to the extra operations introduced, but the cost is still affordable. LGM's inference speed is much faster because of the advantages of 3DGS. However, it is quite hard to extract mesh with 3DGS as representation. Although our proposed method introduces some overhead in the geometry-aware operations, we point out that this overhead is not the bottleneck of the inference performance, but brings large improvement in the generation quality. All inference speed results are tested on a single NVIDIA H800 GPU.

\subsection{More Visualization Results}

We show more of our visualization results of both multi-view generation and single-view reconstruction(Figure \ref{fig:mv_sv_result}). At the top of Figure \ref{fig:mv_sv_result}, we display the 6 input views to our base model, 3 generated high-fidelity novel views and the generated mesh to visualize the results of our multi-view reconstruction. To examine the single-view generation ability of our model, we show some of the results at the bottom of Figure \ref{fig:mv_sv_result}. We collect some images from the internet and use Zero123++ to generate multi-view images of the input, then feed the multi-view images to our M-LRM. The high-quality results demonstrate the superiority of our proposed method.

\section{Conclusion and Limitation}
\noindent \textbf{Conclusion.}
In this work, we introduce the Multi-view Large Reconstruction Model (M-LRM), which is designed to reconstruct high-quality 3D shapes from multiple views in a 3D-aware manner. Our proposed approach incorporates a multi-view consistent cross-attention scheme, enabling M-LRM to accurately extract relevant information from the input images. Additionally, we leverage the 3D priors of the input multi-view images to initialize the triplane tokens. By doing so, our model surpasses the capabilities of instant3D and produces 3D shapes with exceptional fidelity. Experimental results demonstrate that our model substantially improves performance and exhibits faster training convergence compared to instant3D and other LRM-series work.

\noindent \textbf{Limitations and Future Work.}
Although M-LRM has showcased promising performance in generating 3D objects, there are still some limitations that the current framework does not fully address. (i) Currently, our M-LRM assumes the target object is well-defined. Targets of larger scales such as rooms and open-door scenes are still challenging to reconstruct. (ii) Although our model is trained with augmented data, its performance would still inevitably be affected by inconsistent multi-view input when doing single-view to 3D generation tasks. However, we believe that this problem can be solved as the multi-view generation models advance.

\bibliographystyle{IEEEtran}
\bibliography{main}



\end{document}